\documentclass[letterpaper, 10 pt, conference]{ieeeconf}  

\IEEEoverridecommandlockouts                              

\usepackage{times}

\usepackage[utf8]{inputenc}
\usepackage{mathptmx}

\usepackage{comment}
\usepackage{graphicx}
\usepackage{amsmath,amssymb,amsfonts, nccmath} 
\usepackage{booktabs}
\usepackage{bm}
\usepackage{tabularx} 
\usepackage{hyperref}
\usepackage{color}

\title{\LARGE \bf
RoboCAP: Robotic Classification and Precision Pouring of Diverse Liquids and Granular Media with Capacitive Sensing
}
\author{Yexin Hu$^{*}$, Alexandra Gillespie$^{*}$, Akhil Padmanabha, Kavya Puthuveetil, Wesley Lewis, Karan Khokar and\\ Zackory Erickson
\thanks{* Equal Contribution}
\thanks{Yexin Hu, Akhil Padmanabha, Kavya Puthuveetil, Karan Khokar, and Zackory Erickson are with the Robotics Institute,  Carnegie Mellon University, Emails: \{yexinh, akhilpad, kavyap, kkhokar, zerickso\}@andrew.cmu.edu}%
\thanks{Alexandra Gillespie is with Department of Computer Science, Colby College, Email: aggill25@colby.edu}
\thanks{Wesley Lewis is with Department of Computer Science, University of Virginia, Email: wjl4jj@virginia.edu}
}

\begin{document}

\maketitle

\begin{abstract}

Liquids and granular media are pervasive throughout human environments, yet remain challenging for robots to sense and manipulate precisely. In this work, we present a systematic approach at integrating capacitive sensing within robotic end effectors, enabling robust sensing and precise manipulation of liquids and granular media. We introduce the parallel-jaw RoboCAP Gripper with embedded capacitive sensing arrays that enable a robot to directly sense the materials and dynamics of liquids inside of diverse containers. When coupled with model-based control, we demonstrate that the proposed system enables a robotic manipulator to achieve state-of-the-art precision pouring accuracy for a range of substances with varying dynamics properties. Code, designs, and build details are available on the project website\footnote{\href{https://sites.google.com/view/capsense/home}{https://sites.google.com/view/capsense/home}}.

\end{abstract}

\vspace{-0.1cm}
\section{INTRODUCTION}

Identifying and manipulating liquid and granular media, generally held in containers, are fundamental capabilities for robots in settings such as laboratories, factories, and households. A variety of techniques, including visual, haptic, audio, and spectroscopic methods, have been proposed for substance classification and manipulation tasks. However, these methods are challenged by opaque containers or visually indistinguishable substances (e.g. water and vinegar)~\cite{guler2014container, taunyazov2020event}. Some approaches require contact/movement of the containers before classification~\cite{guler2014container, taunyazov2020event, matl2019haptic, huang2022understanding, saal2010active}, or provide limited feedback for control~\cite{hanson2023slurp}. Most prior work on liquid manipulation relies on visual sensing \cite{schenck2017visual, do2019accurate}, and limits coloration of container and substance~\cite{narasimhan2022self}. Furthermore, most methods are only evaluated on a single liquid, generally water \cite{schenck2017visual}, and are not demonstrated for granular media or diverse liquids. 

\begin{figure}[t]
    \centering
    {\includegraphics[width=\linewidth]{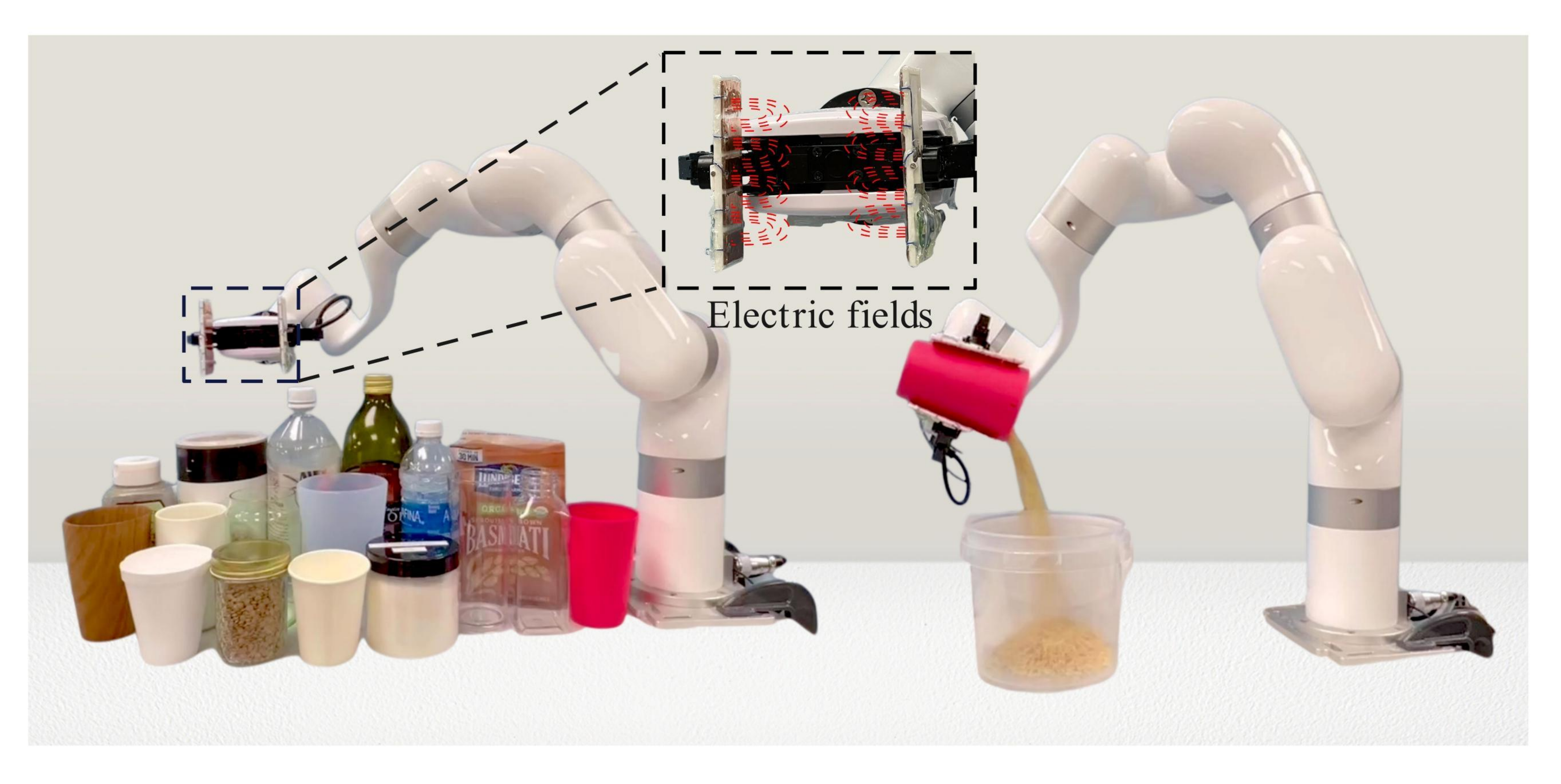}}
    \vspace{-0.8cm}
    \caption{Our capacitive sensing RoboCAP Gripper is mounted on an xArm 7; the highlighting shows two sensing arrays and their electric fields.
    }
    \label{fig:intro}
    \vspace{-15pt}
\end{figure}

In this work, we show how capacitive sensing enables the classification and manipulation of a variety of liquid and granular media, seen in Fig.~\ref{fig:intro}, such that optical properties of the container and its substances are irrelevant. Physical properties of different substances necessitate different manipulation techniques. We develop a set of capacitive sensing arrays, each with 5 electrodes, that can be mounted to the parallel jaws of a robotic gripper. In this case, the capacitance measured by each electrode is directly impacted by the container and the internal substance. Closing the RoboCAP Gripper around different substance-container combinations produces unique variations in capacitance readings such that containers and their substances can be classified with a data-driven approach.

The sensing arrays capture changes in capacitance as media in containers move during pouring. For a set of liquid and granular substances, our model maps changes in capacitance to the weight (grams) poured out of a containerat time $t$, and the substance poured due to inertia after the robot completes its action. From these, we devise a controller that can pour out precise quantities of a substrate from a container with greater precision than a current baseline from literature.

Through this work, we make the following contributions:
\begin{itemize}
    \item We design a robotic end effector, RoboCAP, for liquid and granular media identification and precision pouring with an array of 5 capacitive sensing electrodes integrated on each fingertip.
    \item We demonstrate that capacitive sensing integrated at a robot's end effector allows a robot to accurately infer the materials of household containers, as well as infer the internal liquid or granular media inside of visually opaque containers.
    \item We introduce a model-based controller that enables precision pouring of substances from containers, which we evaluate on a set of five liquid and granular substrates.
\end{itemize}
\vspace{-0.2cm}
\section{Related Work}
Capacitive sensors have had wide ranging industrial applications such as measuring fluid level~\cite{kumar2014review}, proximity, position, humidity and tilt~\cite{terzic2012capacitive}. Recent advances in capacitive sensors include applications such as artificial skins, physical HRI~\cite{erickson2022characterizing}, and in prostheses~\cite{cheng2023recent}, as well as compliant designs for elasticity and flexibility~\cite{Huang2023114500, wu2024proximity,chiurazzi2020novel, yang2021flexible}.

\subsection{Substance and Container Classification} \label{sec:classification_rw}
Tactile sensors have been used with physics-based models to classify fluids within a known container~\cite{matl2019haptic, huang2022understanding, saal2010active}, and with machine learning models for solid media within containers~\cite{chen2016learning, wang2023stev}. The solid media was shaken to produce vibrations, which were measured and utilized by the model, in~\cite{chen2016learning} and~\cite{wang2023stev}. Audio signals (measured either as contact or air pressure vibrations) allows estimation and classification of the substance within a container~\cite{donaher2021audio, clarke2018learning,iashin2021top, eppe2018deep}. More recently, \cite{hanson2023slurp} demonstrated the use of visible and near-infrared (VNIR) spectroscopy embedded within a robot's end effector to classify the material of a container and the internal liquid or granular media substances. We coule capacitive sensing with model-based learning to sense the dynamics of liquids and granular media, allowing for their manipulation.

Multimodal strategies have also been deployed to classify substances within a container. Notable examples include robotic manipulators equipped with vision and tactile sensing~\cite{guler2014container, taunyazov2020event} or vision and audio sensors~\cite{liu2021va2mass, xompero2022corsmal, wilson2019analyzing} to infer substance types and fill levels of liquids inside a container. Notably, classification performance improves substantially for multimodal strategies over individual modalities ~\cite{wilson2019analyzing, liang2019making}. In contrast to tactile and visual sensors, electric fields generated by a capacitive sensor are agnostic to visual occlusions, tactile vibrations, and background audio. Capacitive sensing can achieve competitive material classification and sense liquid and granular media dynamics.

\subsection{Precision Pouring}


Visual methods have frequently been demonstrated for filling a vessel to a target fill height~\cite{do2019accurate, narasimhan2022self} or for pouring out a specific quantity from a container~\cite{schenck2017visual, dong2019precision, zhang2022explainable}. When occlusions prevent visual techniques, works leveraging audio ~\cite{clarke2018learning, liang2019making} and haptic ~\cite{saal2010active, rozo2013force, matl2019haptic} sensing have also been proposed, but with burdens of external signal noise. To overcome the limitations of any individual sensor, multi-modal approaches of  audition and haptics~\cite{liang2020robust}, audition and vision~\cite{wang2023robot}, vision and weight measurements~\cite{kennedy2019autonomous}, or even audition, haptics, and vision~\cite{li2022see}. By contrast, our work shows capacitive sensing, as an independent modality, can achieve strong performance in sensing liquid dynamics through containers, allowing us to pour precise quantities of liquid and granular substances.

Each of these perception strategies has been paired with a variety of control approaches including PID control~\cite{do2019accurate, schenck2017visual}, bang-bang control~\cite{narasimhan2022self}, model-based control~\cite{clarke2018learning}, reinforcement learning~\cite{babaians2022pournet}, and imitation learning~\cite{rozo2013force, zhang2022explainable, li2022see}. In this work, we propose a model-based controller and compare its performance to an imitation learning baseline for pouring using capacitive sensing.

\begin{figure}[t]
    \centering
    {\includegraphics[width=\linewidth]{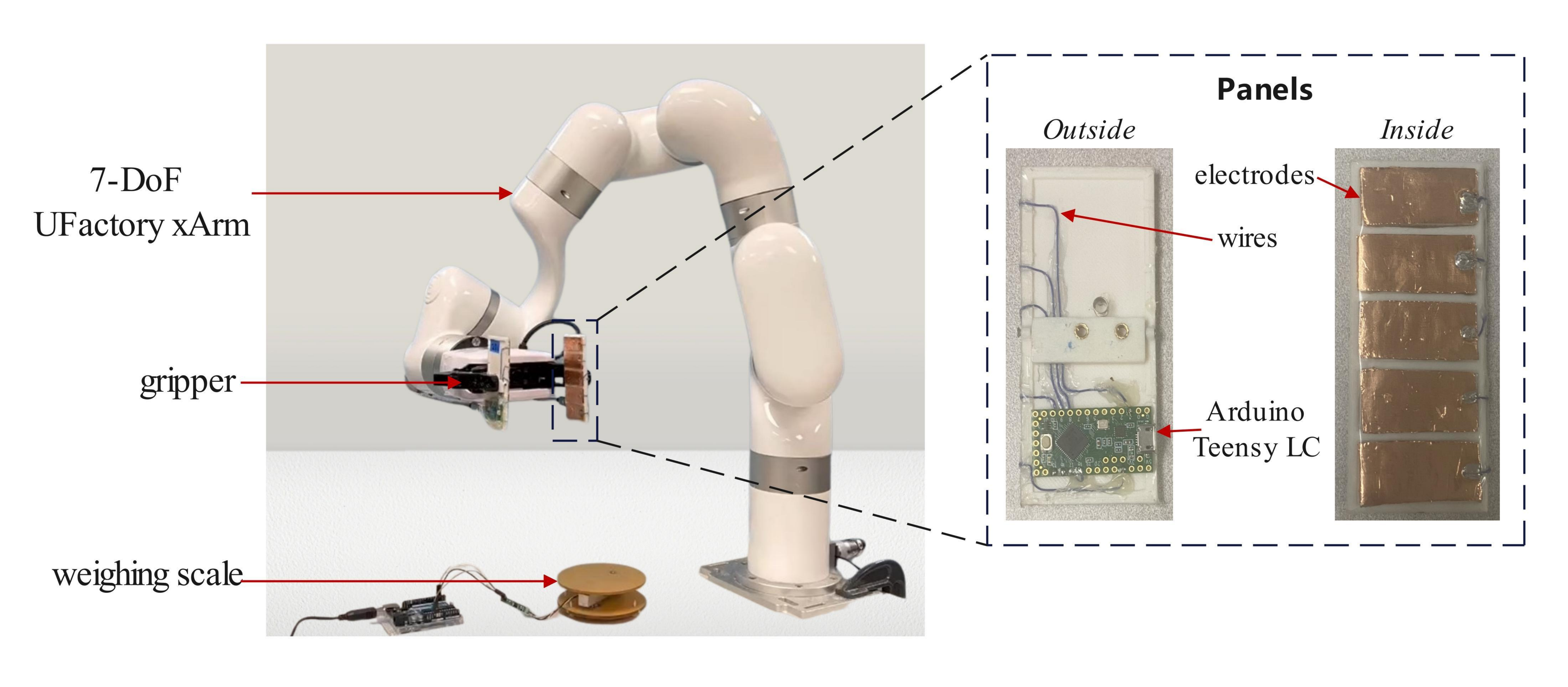}}
    \vspace{-0.8cm}
    \caption{Two 3D-printed panels, each consisting of five copper electrodes and a Teensy LC, are shown mounted on the  UFactory xArm gripper. The experimental setup is shown with a weighing scale to measure ground truth for robotic pouring experiments.}
    \label{fig:sensor}
    \vspace{-15pt}
\end{figure}
\vspace{-0.2cm}
\section{Sensor Design}
\label{sec:sensor_design}
As shown in Fig.~\ref{fig:sensor}, we designed a multi-electrode capacitive sensing system integrated within the parallel-jaw gripper of a UFactory xArm7 robotic manipulator. Capacitance is the capability of a material to hold electric charge. Our RoboCAP system measures the self-capacitance of the container and substance directly in the vicinity of each electrode. We designed two 3D-printed panels, each with an array of five copper foil electrodes, and attached them to both jaws of the RoboCAP Gripper, with a small spring to provide compliance and adherence to the shape of non-cylindrical containers. In order to improve signal quality and diversity between electrodes, particularly during robotic pouring, as seen in Fig.~\ref{fig:intro}, the capacitive arrays are mounted vertically to the panels to increase sensor distribution along the height axis of a container. The electrodes on each panel are connected to a Teensy LC microcontroller, mounted on the outside of the panel. The Teensy LC has dedicated hardware pins which use a 1-pF internal reference capacitor to account for environmental factors that affect capacitance readings such as temperature and humidity. We record data from all 10 electrodes at a frequency of 100~Hz. A thin layer of waterproof polypropylene tape protects the electrodes on the inside of the panels. A plastic film and hot glue are used to waterproof the electronics on the outside of each panel. From preliminary experiments, we determined the sensing range to be 20 mm, with high sensitivity in the 10 mm range. Our models' temporal architecture allows the model to analyze the dynamic changes in capacitance signal over time, enabling the inference of different materials and estimate pour quantities without prior calibration.
\vspace{-0.2cm}
\section{Substance and Container Classification}
\label{sec:classification}
{
\subsection{Classification Dataset}
\label{sec:classificationdataset}
We collected a classification dataset using nine commonly found household containers and nine substances, several of which are found in other studies investigating classification of liquids or granular media~\cite{hanson2023slurp}. The containers used were paper, styrofoam, ceramic, glass, wood, silicon, polyethylene terephthalate (PET) plastic, polypropylene (PP) plastic and polycarbonate (PC) plastic. The substances used were oats, vinegar, oil, honey, starch, rice, lentils, sugar and water. All containers were opaque except glass, PET, and PC, and all were non-cylindrical, i.e. shaped as a frustum of a cone, except ceramic and glass. The liquid and granular media we used had varying viscosity and opacity.


Capacitance data was collected for every combination of container and substance, resulting in 81 total classes. To account for any marginal variations from environmental factors, we collected data on 4 different days at varying times. The open position for the RoboCap gripper is 90 mm wider than the closed position for each container. To collect each dataset, we captured capacitance measurements for 10 closing  iterations of the RoboCAP Gripper, for each one of the 81 classes. 

It takes the RoboCAP Gripper 2 seconds to traverse from the open to the close position, during which time the capacitance data was collected continuously. This resulted in a total of 200 individual sensor measurements per iteration of closing for each of the 10 electrodes, due to the sampling rate of 100~Hz. Sample plots of capacitance during data collection for water and oil substances in glass and PP plastic containers are shown in Fig. \ref{fig3:capacitanceCharacterization}.

\begin{figure}[t]
    \rotatebox{270}{\includegraphics[scale=.31]{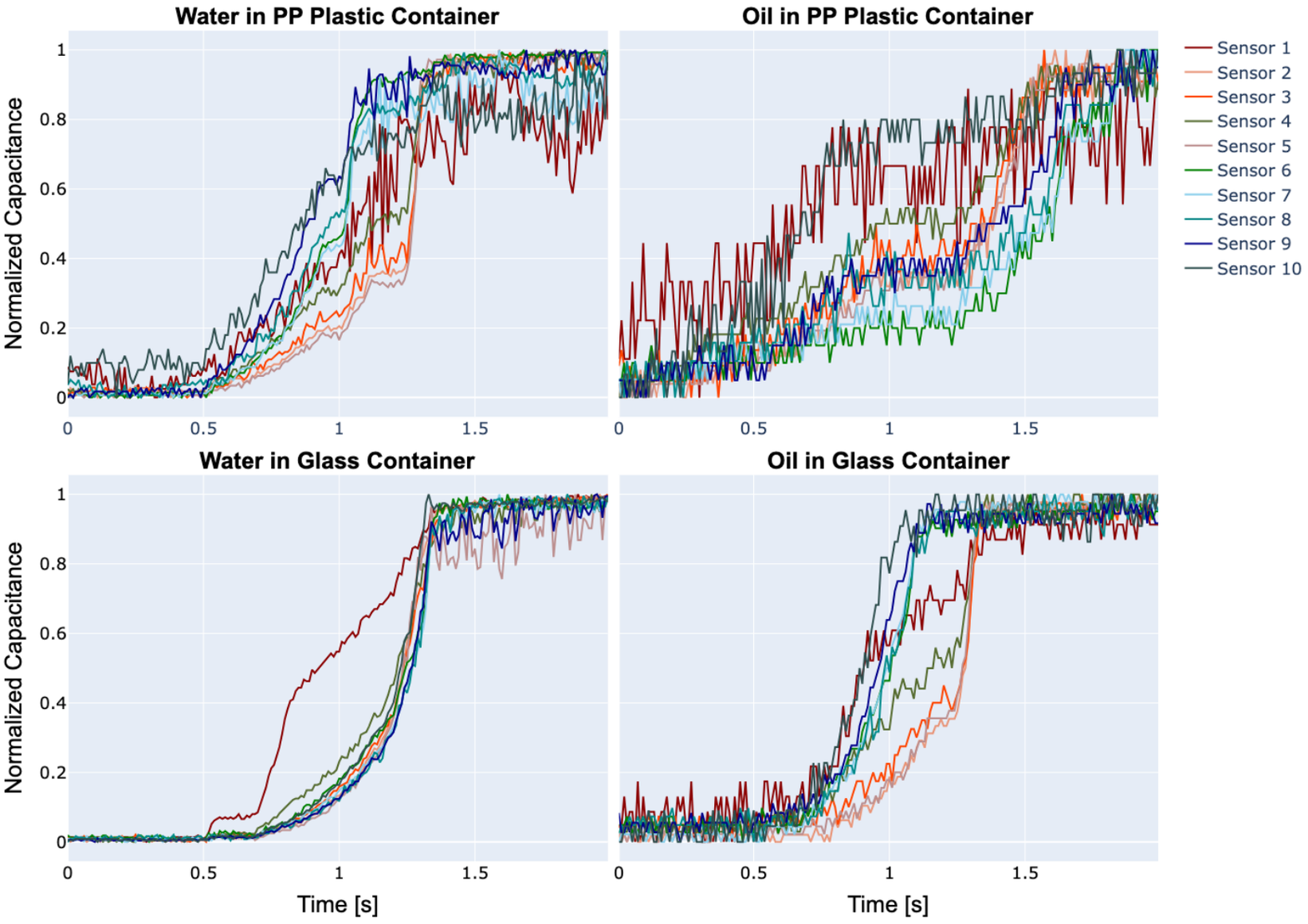
}}
    \vspace{-0.6cm}
    \caption{Capacitance data over time is shown during a grasping cycle of PP plastic and glass containers, containing water and oil. As the RoboCAP Gripper closes around the container, the sensor values increase. Each line on the plot is min-max normalized and corresponds to one of ten electrodes on the sensing arrays.}
    \label{fig3:capacitanceCharacterization}
    \vspace{-15pt}
\end{figure}

\subsection{Feature Extraction and Model Training}
We partition our data into 2 second windows, corresponding to a single RoboCAP Gripper closing action. In each window of data, for each electrode, $e$, the measured capacitance values, $c_{e}\in\mathbb{R}^{200}$, are used as input features. We concatenate the capacitance values for all 10 electrodes to obtain our input feature vector, $f = [c_{1}, c_{2}, ..., c_{10}] \in\mathbb{R}^{2000}$}. In total, there are 3240 data samples (10 iterations per substance-container pair * 4 days * 81 classes).

We use two different methods to split the dataset into training and test for evaluation, (1) We randomly take 75\% of them as training and validation dataset, 25\% as test dataset (with a same number of samples for each substance-container pair); (2) We take the first 3 days data as training and validation dataset, the remaining day as test dataset. We then standardize the training features (zero mean and unit standard deviation) and apply the computed fit to the test set. 

We use the residual multi-layer perceptron~\cite{haykin1998neural, he2016deep} as the main architecture of our classifier. The model first maps input features to a hidden representation using a 256-node fully connected layer with a ReLU activation function. The transformed features are then processed through a stack of 4 residual blocks, each consisting of a 256-node fully connected layer with ReLU activation and a residual connection followed by ReLU to preserve information flow. Finally, the output is mapped to the number of classes through a fully connected layer, producing logits for classification. We train our model using the Adam Optimizer, with full-batch training, a learning rate of $10^{-3}$, weight decay of $10^{-2}$ and Cross Entropy Loss as the loss function. The results are described in Section~\ref{sec:results}.

\begin{figure*}[t]
    \centering
    {\includegraphics[width=0.95\linewidth]{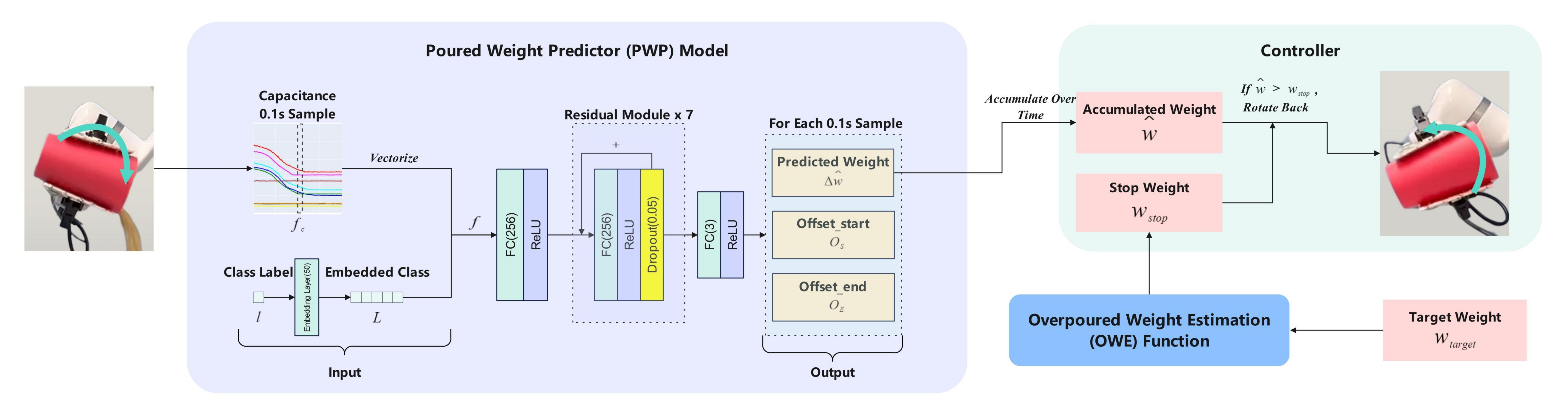}}
    \vspace{-0.5cm}
    \caption{RoboCAP Controller pipeline. The Poured Weight Predictor (PWP) model takes 0.1~s windows of normalized capacitance signals and the substance class as input; the outputs are the estimated change in weight per 0.1~s, $\Delta\hat{w}$, and two offset terms, $O_S$, and $O_E$. The Overpoured Weight Estimation (OWE) function takes the target weight $w_{target}$ as input and calculates the stop weight $w_{stop}$. Our controller accumulates the predicted $\Delta\hat{w}$ during the pouring process and compares the accumulated weight $\hat{w}$ with $w_{stop}$. Once the accumulated weight $\hat{w}$ exceeds $w_{stop}$, our controller starts to rotate back.}
    \label{fig:model}
    \vspace{-.3cm}
\end{figure*}

\begin{figure*}[t]
    \centering
    {\includegraphics[width=0.95\linewidth]{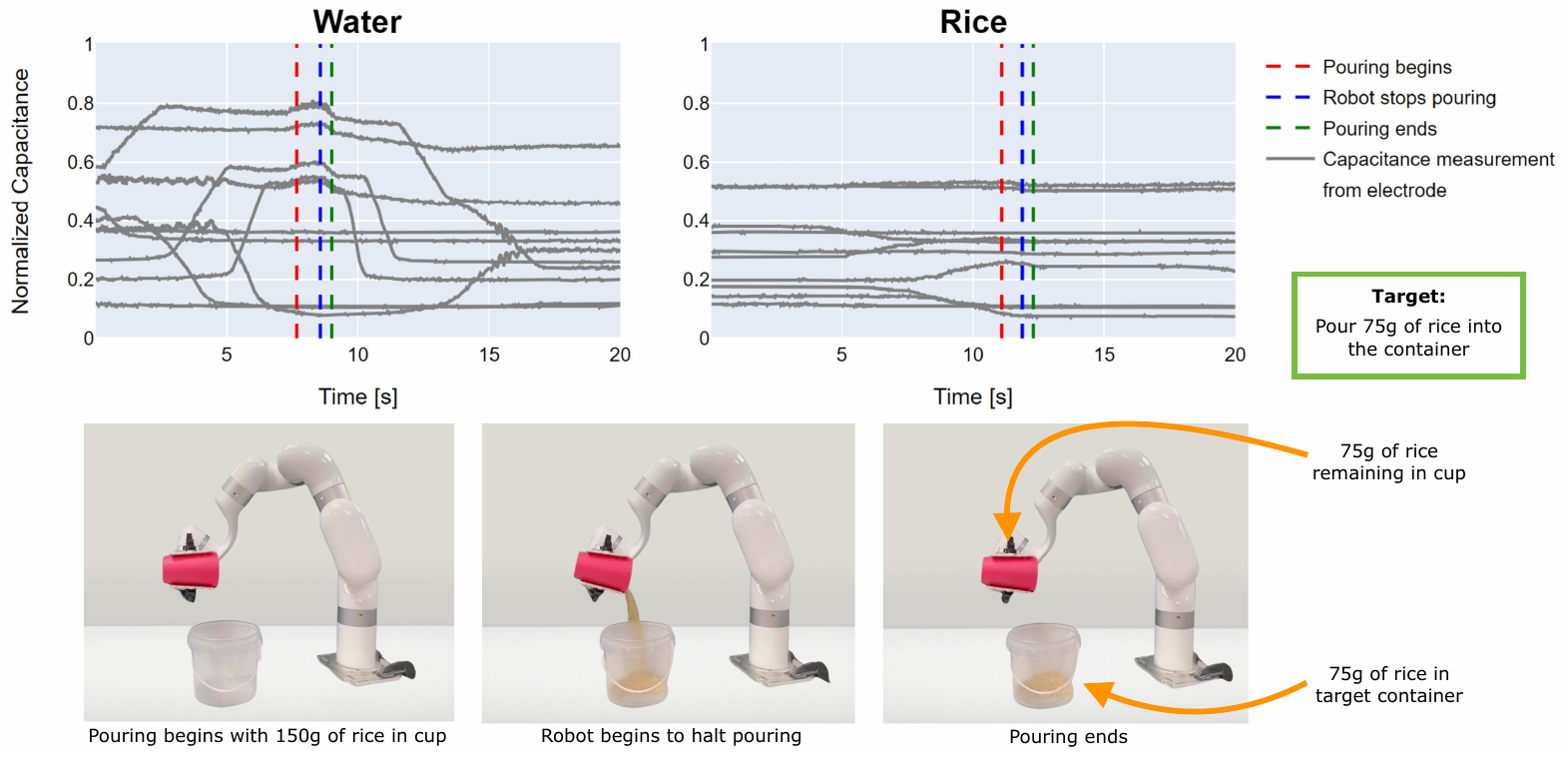}}
    \vspace{-0.45cm}
    \caption{The upper plots show the normalized capacitance over time for water and rice during the pouring process with a target weight of 75g. The red dashed line indicates the moment when the arm starts to pour the substance from the container. The blue dashed line indicates the moment when pouring stops and the RoboCAP Gripper starts to rotate backward. The green dashed line indicates the moment when the substance stops being poured out from the container. The three images below the graph correspond to the experimental pouring images of rice at these three stages.}
    \label{fig:cap}
    \vspace{-18pt}
\end{figure*}
\vspace{-0.2cm}

\section{Pouring Estimation and Control}
\label{sec:pouringmethods}
\subsection{Control Pipeline}
We propose a method, shown in Fig.~\ref{fig:model}, that allows a robot arm to precisely pour a specific quantity from a container, relying only on the knowledge of the substance being poured and real-time capacitance measurements. To perform precision pouring, a robot must first learn the relationship between the amount of substance poured and the corresponding capacitance measurements at each timestep in the pouring trajectory. The robot must then use this information to determine when to stop pouring. The inertia of the substance as it's being poured makes this task challenging as the manipulator cannot stop pouring instantaneously. To prevent overpouring, a robot must proactively stop pouring \textit{before} the target quantity is reached. Capacitance data and this pouring process is further shown visually in Fig.~\ref{fig:cap}.

To tackle these challenges, we present a two-component approach to robotic precision pouring. The first component, which we call the Poured Weight Predictor (PWP), is a residual multi-layer perceptron~\cite{haykin1998neural, he2016deep}, depicted in Fig.~\ref{fig:model}. It estimates the weight that has been poured, up until the robot starts rotating the container back to the upright state to stop pouring. The second component, called the Overpoured Weight Estimation (OWE) function, is a polynomial function that estimates how much of the substance will be ``overpoured'' while the robot attempts to stop pouring. Putting both pieces together, our controller rotates the RoboCAP Gripper forward and uses the PWP to predict how much of a substance has been cumulatively poured out. When the stop weight estimated by the OWE is reached, the controller starts to rotate the RoboCAP Gripper back so that the precise target weight is poured out by the time the gripper and substance stream come to a complete stop, thus preventing overpouring.

\subsection{Experimental Setup}

\label{sec:pouringdataset}
Due to the time cost of conducting deformable substance experiments, we chose five representative liquid and granular substances: water, vinegar, oil, rice, and lentils. Our objective was to assess the feasibility of generalizing precision pouring across varying substances. Thus, all substances were poured only from a polypropylene (PP) plastic container. Before pouring, each container was filled with 150g of a given substance. The container was grasped such that the top of the RoboCAP Gripper was 1~cm below the upper edge of the container, ensuring all capacitive electrodes contacted the container. The pouring action was a clockwise rotation of the robot's wrist joint at a speed of 9$^{\circ}$/s, as shown in Fig. ~\ref{fig:intro}. The maximum pouring angle for data collection for training was 123.5$^{\circ}$ from the container's vertical orientation, 0$^{\circ}$. We captured capacitance measurements from all the electrodes at 100~Hz. Ground truth weight measurements were captured at 10~Hz from a strain gauge based weighing scale (0-5kg, $\pm$1g) placed beneath the target container.

\vspace{-0.2cm}
\subsection{Poured Weight Predictor (PWP)}
\subsubsection{PWP Model Architecture and Training Details}

Fig.~\ref{fig:model} depicts the neural network architecture for the Poured Weight Predictor (PWP) model. We utilize a residual multi-layer perceptron architecture, which consists of 7 residual modules with a 256-node fully-connected layer, followed by a ReLU activation, and a dropout of 0.05 within each module with the final output layer followed by a ReLU activation. We train using the Adam Optimizer, with a batch size of 128, learning rate of $10^{-4}$, weight decay of $10^{-3}$.

\subsubsection{PWP Dataset}

To collect data to train the model, the robot performed 10 complete pouring trials for each of the five substances described in Sec.~\ref{sec:pouringdataset}, totaling 50 trials. In each trial, the wrist rotates to pour until it reaches the maximum pouring angle. Each pouring trial lasts 20 seconds. The full dataset of 50 total pouring trials is divided such that 80\% (40 trials) are used for training and 20\% (10 trials) are used for validation of the PWP model.

\subsubsection{PWP Model Inputs}
\label{sec:PWPinputs}
First, we partition our capacitance data using a sliding window of size $h=0.1$ seconds, with a stride of 0.01 seconds. This results in 1991 samples per 20-second trial, with a total of 79,640 samples (1991 samples * 40 trials) for training and 19,910 samples (1991 samples * 10 trials) for validation. For each electrode, $e$, within a window of data, we have measured capacitance values, $c_{e}\in\mathbb{R}^{10}$. The data for all 10 electrodes is concatenated to form a vector, $f_c = [c_{1},  c_{2},...,c_{10}] \in\mathbb{R}^{100}$. Min-max normalization is applied to scale all capacitance values such that $f_c \in [0, 1]$. The model also receives as input the ground truth substance class label, represented as an integer $l\in [0, 1, \ldots, 4]$. This class label is then embedded into a 50-dimensional vector $\mathbf{L}\in\mathbb{R}^{50}$. The embedding layer is learned through backpropagation with the rest of the neural network model. $\mathbf{L}$ is concatenated with $f_c$ to form the input features, $f\in \mathbb{R}^{150}$, for the PWP model.

\subsubsection{PWP Model Outputs}
The first output of the PWP model is the predicted change in weight from the pouring container, $\Delta\hat{w}$, for each window. The predicted total weight, $\hat{w}$, at each timestep $t$, is equal to the cumulative sum of $\Delta\hat{w}$ for all windows from the start of the pouring trajectory up to the current timestep $t_0:t$. The cumulative ground truth weight poured from the container, $w$, at each timestep sampled from the scale at 10~Hz, is linearly interpolated to match the frequency of capacitance measurements.

Since we cannot directly measure the total weight of the substance as it leaves the \textit{pouring container}, we instead must measure the total weight in the \textit{target container} using a scale. Given the impulse induced by a falling substance will cause a certain amount of deviation between the reading of the weighing scale and the actual value, (1) we use a low pass filter to smooth the training weight data; (2) due to the momentum of falling substances, the measured weight will typically decrease slightly at the end of a pour (see Fig. \ref{fig:weight}); and we remove the effect of momentum by setting $w_{t-1s}$ $ \leftarrow $ $w_t$ whenever the measured weight has decreased at time $t$ (i.e. $w_t$ $<$ $w_{t-1s}$). We set a 1 second threshold to account for only momentum and not noise in the scale. The time delay between when the substance leaves the lip of the pouring container and when it reaches the target container introduces some ambiguity in computing the ground truth change in weight from the pouring container $\Delta{w}$, at a given timestep $t$. To account for this delay, which is different for each substance and is also dependent on the distance between the lip of the container and the scale, we include two learned outputs in the PWP model, offset\_start, $O_S$, and offset\_end, $O_E$, for each $h=0.1$s window of data. $O_S$ accounts for the time delay between the weight measurements and capacitance signals at the beginning of each window while $O_E$ accounts for the time delay at the end. Thus, we can define the change in the ground truth weight at a particular time $t$ as $\Delta w = w_{t + h + O_E} - w_{t + O_S}$.
\subsubsection{Loss Function and Model Training}
\label{sec:loss}
We define the training objective for the PWP as the sum of four component loss terms:
\vspace{-.3cm}
\begin{equation}\label{eq:full_loss}
L = L_{weight} + {\alpha}L_{aux} + {\beta}L_{p1} + {\gamma}L_{p2}
\vspace{-.15cm}
\end{equation}
where $\alpha$, $\beta$ and $\gamma$ are hyperparameters for balancing these four losses. We set $\alpha=0.2$, $\beta=1$, and $\gamma =0.1$ based on preliminary experiments.

The primary component loss term minimizes the differences between the $\Delta\hat{w}$ and the changes in ground truth weight, $\Delta{w}$, at corresponding timesteps, allowing the PWP to estimate the poured-out weight, $\Delta\hat{w}$, at every time window $h$. This loss term is given by:
\begin{equation}\label{eq:loss_1}
L_{weight} = \lVert \Delta\hat{w} - \Delta w \rVert^2_2
\vspace{-.15cm}
\end{equation}

The second component loss term, $L_{aux}$, acts as a strong constraint on \eqref{eq:loss_1} to prevent \eqref{eq:loss_1} from being exploited during gradient descent and producing zero values for all the predicted weights. This loss term ensures that at any given timestep, $t_{rand}$, the sum of the predicted weights $\Delta\hat{w}$ in all previous $h=0.1$s time windows, should be equal to the ground truth weight at that timestep, $w_{t_{rand}}$. The timestep $t_{rand}$ is randomly selected for each iteration during the training process. We specifically choose $t_{rand}$ to be randomly selected rather than calculated for all time steps to maintain computational efficiency.

\begin{equation}\label{eq:loss_2}
L_{aux} =\left\lVert \left({\sum_{t=0}^{t_{rand} - h}{\Delta \hat{w}_{t:(t+h)}}}\right) - w_{t_{rand} + O_E}\right\rVert^2_2
\end{equation}

To ensure (\ref{eq:loss_2}) is a differentiable operation, we first round float time indices $t_f$ to its upper integer bound $t_u$ and lower integer bound $t_l$, and then linearly interpolate between $w_{t_l}$ and $w_{t_u}$ to obtain the weight at $t_f$, for example, to obtain the weight at t = 100.2 (i.e., 1.002 seconds), we interpolate between $w_{100}$ and $w_{101}$. This interpolation allows the gradient to flow through backpropagation.

To ensure that $O_S$ and $O_E$ are not artificially low values, we also add two penalty losses, $L_{p1}$ and $L_{p2}$, as follows:

\begin{multline}\label{eq:loss_3}
L_{p1} = \max(0, O_{min} - O_S) + \max(0, O_{min} - O_E)
\end{multline}

\begin{equation}\label{eq:loss_4}
L_{p2} = \left\lVert {O_S}_{t:(t + h)} - {O_E}_{(t - h):t} \right\rVert^2_2
\vspace{-.15cm}
\end{equation}

Based on a frame-by-frame analysis of pouring videos for each substance, the minimum delay $O_{min}$ between when the poured substance initially leaves the container and eventually contacts the collecting cup is 0.15s. Since it is physically impossible for $O_S$ and $O_E$ to be shorter than this minimum delay, we define $L_{p1}$ to penalize $O_S$ and $O_E$ that are less than $O_{min}=0.15$s during training. Here we apply a soft constraint to allow flexibility in the model's predictions, especially for predicting $O_S$ and $O_E$ during the period before the starting of pouring and after the ending of pouring, when there is no substance poured out. Additionally, we reason that the time it takes for the substance to reach the collecting cup should not vary significantly between adjacent timesteps. Thus, $L_{p2}$ is used to penalize the cases where there is a large discrepancy between the $O_E$ predicted from one window and the $O_S$ predicted from the following window.

\subsection{Overpoured Weight Estimation (OWE) Function}
\label{sec:control_overpouring}
Our PWP model can be used to command the RoboCAP Gripper to stop pouring once the predicted cumulative weight reaches some target. However, after the pouring action stops, some of the substance will still pour out of the container due to inertia, surpassing the target weight in the receiving container. Thus, we must determine the weight at which the robot should begin rotating its RoboCAP Gripper to stop pouring, $w_{stop}$, to avoid exceeding the target weight $w_{target}$. We define, $w_{stop}$ as follows, 

\vspace{-.3cm}

\begin{equation}\label{eq:loss_7}
w_{stop} = w_{target} - w_{overpoured}
\end{equation}

\vspace{-.2cm}
$w_{overpoured}$ generally increases with $w_{stop}$ but is also influenced by environmental noise factors. To balance complexity and generalizability, we chose to approximate the relationship between these two variables using a second-degree polynomial. We avoided a linear model as the relationship is not linear, and we opted against higher-degree polynomials to prevent overfitting to the environmental noise. We call this second-degree polynomial Overpoured Weight Estimation (OWE) function, denoted as: 
\vspace{-.3cm}
\begin{equation}\label{eq:loss_6}
w_{overpoured} = aw_{stop}^2 + bw_{stop} + c
\vspace{-.15cm}
\end{equation}

where $a$, $b$, and $c$ are the coefficients of the second-degree polynomial which vary with the substance type.

We collect data to fit the OWE by executing 24 additional pouring trials per substance. In each trial we use the trained PWP model to pour out target quantities of each substance in 10g increments ranging from 30g to 100g, repeated 3 times per quantity. Since the PWP alone cannot account for overpouring, the robot will only rotate the container back when it predicts that the target weight, $w_{target}$, has been reached, causing the substance to inevitably be overpoured. The difference between the weight that the PWP model predicts and the ground truth weight is the overpoured weight, $w_{overpoured}$. We use ordinary least squares to fit the second-degree OWE polynomial to the collected data. 

Now, we can use PWP to estimate the total poured weight using the cumulative sum of $\Delta\hat{w}$, and the OWE model to calculate $w_{stop}$ so that overpouring doesn't occur. To evaluate our approach, we use the trained PWP model and fitted OWE function for control to pour each of the five substances at four different target weights (50g, 75g, 100g, and 125g) ten times. The results from this evaluation is described in Section~\ref{sec:results}.

\vspace{-.2cm}
\subsection{Behavior Cloning Baseline}

Behavior cloning (BC) is a commonly employed method in prior robotic pouring literature~\cite{zhang2022explainable, li2022see}, as well as in robotic manipulation more broadly~\cite{zhang2018deep}. We implement BC by rotating the robot gripper following the same method as described in Sec.~\ref{sec:pouringdataset} and then stopping and rotating back at a manually specified time to reach the target weight. We save only demonstrations that have less than 2g of error from the target pour weight. The BC model takes as input the concatenated feature vector provided to the PWP model, $f$, as well as the target weight class label (50g, 75g, 100g, or 125g). The BC model shares a similar architecture as that shown in Fig.~\ref{fig:model}, with a 1-dimensional output and sigmoid activation that indicates the direction of end effector rotation for pouring. We train using binary cross entropy loss and all other hyperparameters matching PWP. The behavior cloning model reaches its minimum validation loss within 50 epochs. 

To train this BC model, we collect five demonstrations for each substance at each target weight (in total 100 demonstrations), with a training and validation split of 80/20. To evaluate the trained BC model, we conduct 10 test trials for each substance and target weight.

\vspace{-.1cm}
\section{Results and Discussions}
\label{sec:results}
{

\subsection{Substance and Container Classification} {
We use SLURP~\cite{hanson2023slurp} as the baseline to compare. SLURP trains separate MLPs simultaneously to predict containers and substances separately. Each MLP consists of 3 fully connected layers of 200, 100, 25 neurons with ReLU activation followed. This model takes the concatenation of capacitance readings and the gradient of the signal [$f$, $f^{\prime}$] as input, and passes it into the container MLP to predict the container label. The prediction of the container is then concatenated with the original input, to form a new input that is passed through the substance MLP to predict the substance label. We train it using the same parameters as our model.

Table~\ref{table:classification3models} presents the classification results of our RoboCAP classifier and the SLURP baseline over the test set for two different dataset splitting methods as described in Sec.~\ref{sec:classification},  and Fig.~\ref{fig:ConfusionM} shows the confusion matrices of our classifier over test dataset. Our classifier performs better than the baseline model. For the randomly splitting method, our model correctly infers the class labels of containers with 98.6\% accuracy and of substances with 95.8\% accuracy, with 95.6\% accuracy in estimating both container and internal substance correctly for all 81 combinations.  The accuracy of the model using a second splitting method (for generalizability) was 96.4\% for containers, 83.2\%  for the substances, and 82.8\%  for classifying both container and substance correctly. For unseen distributions, the model will confuse very similar capacitance readings for different container-substance combinations (for example, rice and lentils in the same container have near-indistinguishable signals upon visual inspection) but still perform well. 

Overall, these results suggest that end-effector capacitive sensing presents a competitive methodology for inferring the materials of rigid objects and liquids and granular media inside of diverse containers.

\vspace{-.2cm}
\begin{table}[htbp]
\caption{test set classification accuracies for container, substance, and overall for leaving one day out and randomly splitting
}
    \vspace{-.3 cm}
    \centering
    \begin{tabular}{cccc}  \toprule &
     Container& Substance & Overall\\ \midrule\midrule
     RoboCAP (ours) (Leave 1 Day Out) & 96.4\%  & 83.2\%  & \textbf{82.8\%}  \\
     \midrule
      RoboCAP (ours) (Randomly Split)  & 98.6\%  & 95.8\%  & \textbf{95.6}\% \\
     \midrule
      SLURP (Leave 1 Day Out)  & 94.7\%  & 68.5\%  & 65.3\%  \\
     \midrule
      SLURP (Randomly Split)  & 96.1\%  & 81.5\%  & 79.5\% \\
     \bottomrule
    \end{tabular}
    \label{table:classification3models}
    \vspace{-10pt}
\end{table}

\begin{figure}[htbp]
    \centering
    \vspace{-0.34cm}
    \includegraphics[width=\linewidth]{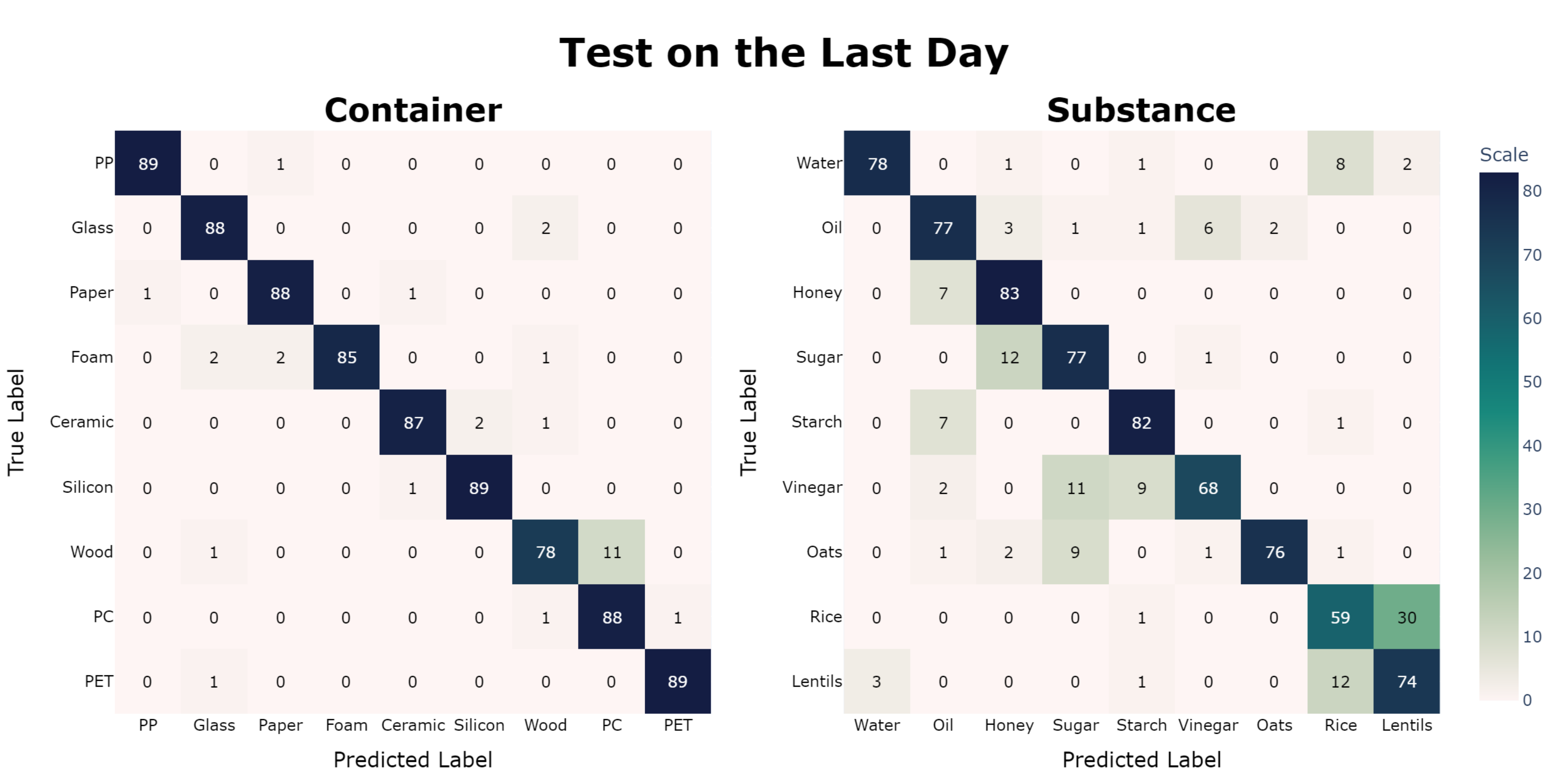}

    \vspace{-0.26cm}

    \includegraphics[width=\linewidth]{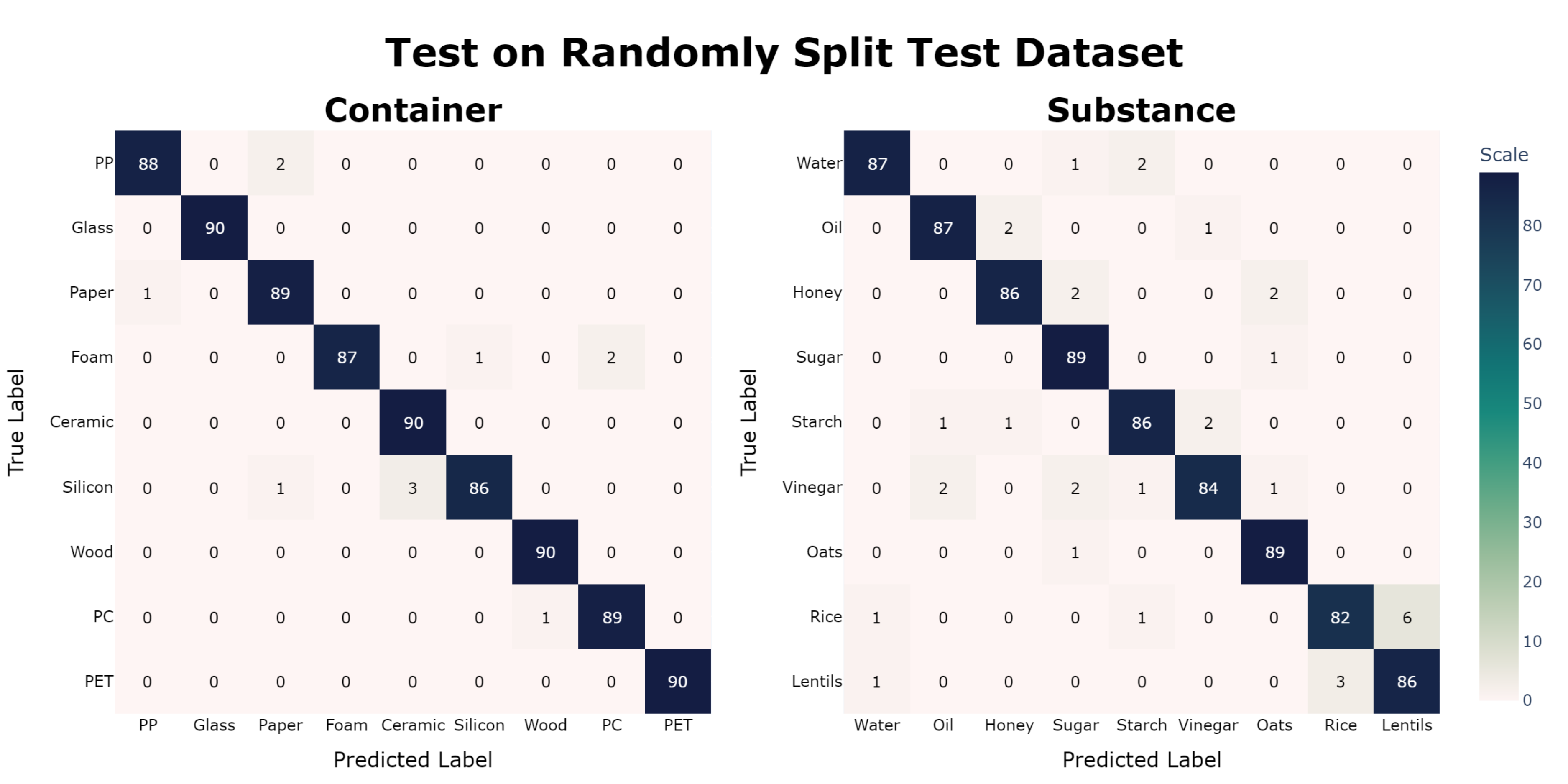}

    \vspace{-0.35cm}
    \caption{Confusion matrices for container and substance classification. the first row is the result of testing on the last day, and the second row is the result of testing on the randomly split test dataset}
    \label{fig:ConfusionM}
    \vspace{-10pt}
\end{figure}

\begin{figure}[t]
    \centering
    {\includegraphics[width=0.95\linewidth]{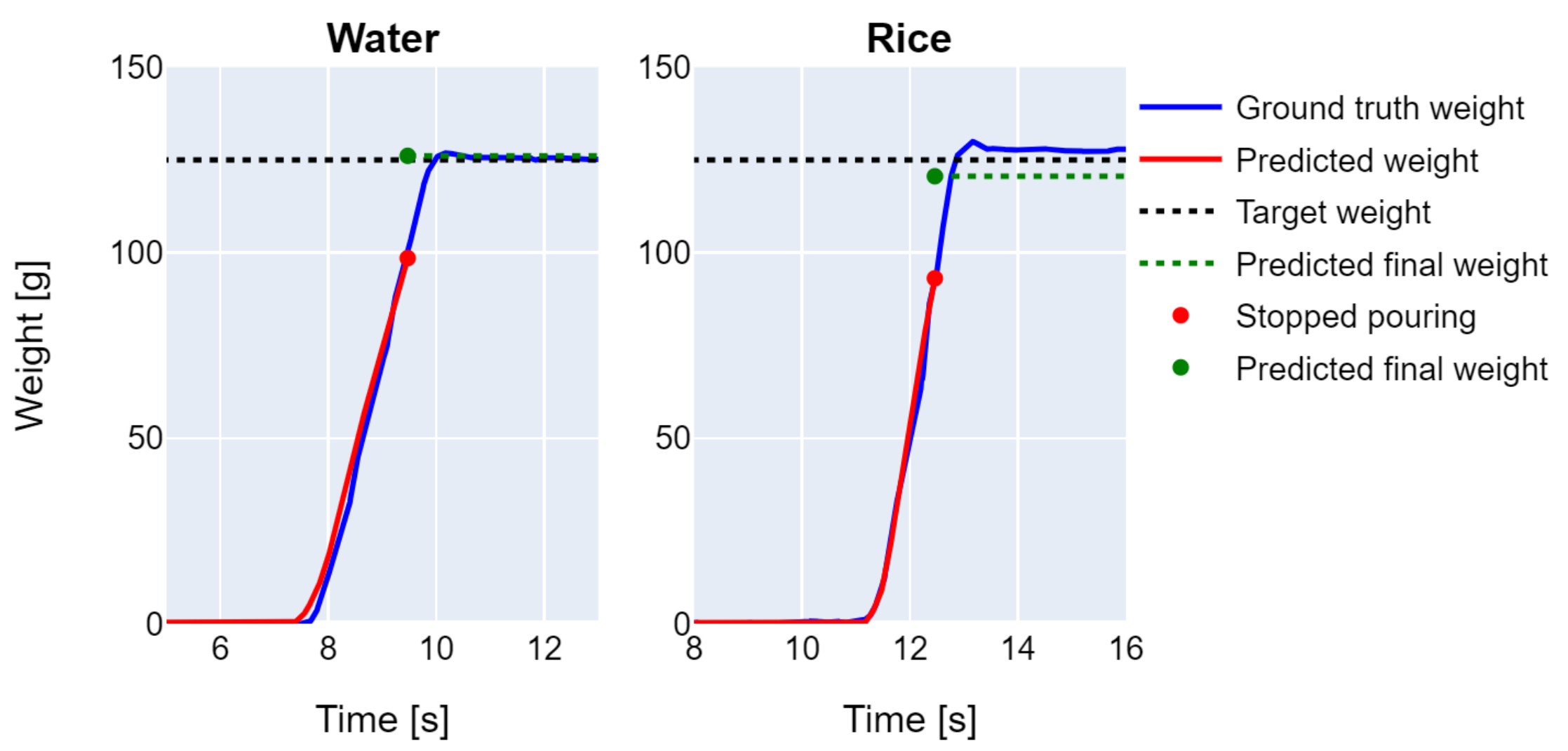}}
    \vspace{-0.3cm}
    \caption{These two sample plots show predicted and ground truth weight changes over time for water and rice at a target weight of 125g. The black dotted line is the target weight, each blue line is the ground truth weight change, and the red line is the predicted weight change. Here, the red lines have been shifted to the right with the predicted offset, $O_E$, to align with the ground truth timeline. The red point indicates the predicted weight when pouring stops, and the green point indicates the predicted final weight.}
    \label{fig:weight}
    \vspace{-10pt}
\end{figure}

}

\begin{figure}
    \centering
    \includegraphics[width=0.9\linewidth]{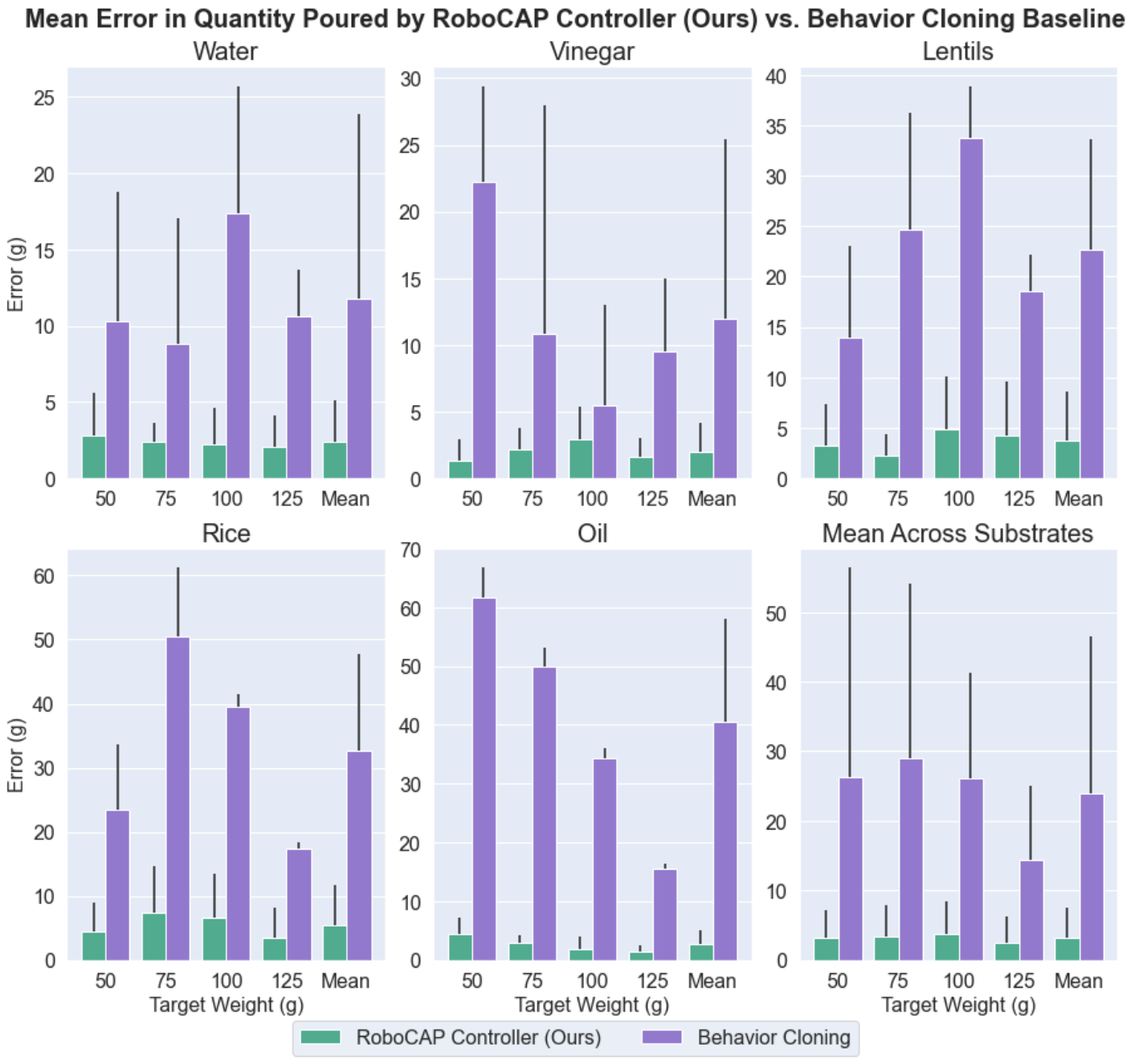}
    \vspace{-0.4cm}
    \caption{RoboCAP Controller's mean error is less than the behavior cloning baseline across all substances and all target weights.}
    \label{fig:mean_error_plots}
    \vspace{-17pt}
\end{figure}

\subsection{Precision Pouring} {
In addition to substance and container classification, capacitive sensing coupled with a model-based control strategy can enable a robot to precisely pour a diverse array of liquids and granular media. Sample plots are demonstrated in Fig.~\ref{fig:weight}, where the red lines represent the amount predicted by our RoboCAP Controller system to have been poured into the receiving container, and the blue line represents the ground truth weight. It can be seen that the red and blue lines coincide during the pouring action, indicating that our RoboCAP Controller system can accurately predict the mass poured for the various substances over time. 


In Fig. \ref{fig:mean_error_plots}, we provide the mean absolute error in grams of our RoboCAP Controller for pouring the five substances with four target weights. We also show the mean results across all five substances for each target weight. Across all substances and target weights, we achieve a low average error of 3.2g.

Our RoboCAP Controller approach's average poured error is below 3g for all three liquids, but we notice that granular media have higher pouring errors in comparison to the liquids. We suspect the friction between individual particles of granular media can cause piling and lead to random, sudden "avalanches" when the end effector pour angle becomes steep enough. This may also be due to larger flow rates for lentils (8.2g/0.1s) and rice (10.2g/0.1s) in comparison to the three liquids (5-6 g/0.1s). Predicting changes in weight at higher frequencies could alleviate this problem. 

In comparison to past works~\cite{schenck2017visual, do2019accurate, narasimhan2022self, kennedy2019autonomous, dong2019precision, wang2023robot, babaians2022pournet}, RoboCAP obtains low error for robotic pouring while also generalizing to a diverse set of granular media and liquids with vastly different physical properties. Notably, the PWP model allows our method to predict when pouring begins in addition to the overall pour rate. In contrast to all prior work, RoboCAP precisely pours by estimating weight of the substance leaving the container allowing us to achieve state-of-the-art performance of 2.3 g mean error over 200 trials with target pour weights ranging from 50g-125g regardless of container opacity.


As seen in Fig.~\ref{fig:mean_error_plots}, we compare our model-based pouring approach to a behavior cloning (BC) baseline, which directly selects the direction of rotation of the RoboCAP Gripper during pouring. The BC approach has an average error of 23.9g over all trialswhile our RoboCAP Controller method achieved an average error of 3.2g.

Behavior cloning can usually only complete tasks with demonstrations; as a result, in our case, where four target weights are demonstrated, it can only complete pouring tasks for these weights while our model-based system works for any target weight within the training distribution.

}
\section{CONCLUSIONS}
\vspace{-0.1cm}
In this work, we show how capacitive sensing is a promising modality for the classification and manipulation of liquids and granular media. We present the RoboCAP Gripper, a robot end effector with embedded capacitive sensing arrays. Capacitive data collected with this end effector can accurately distinguish between 81 unique substance-container classes. Generalization of capacitive end-effector sensing to novel, out-of-distribution containers or substances presents a promising yet challenging direction for future exploration. One approach to doing so is to predict material properties~\cite{boonvisut2012estimation, huang2022understanding, guevara2018stir} rather than relying solely on discrete class labels. We then presented the RoboCAP Controller, which precisely pours specific quantities of five different substrates, while minimizing overpouring, achieving superior pouring performance.
Future work could focus on improving the classification and pouring model architecture, fine-tuning hyperparameters, multimodal strategies with spectroscopy~\cite{hanson2023slurp}, and exploring alternative methods to calculate overpoured weight for greater precision and efficiency, as well as systematically testing precision pouring performance across a wider variety of containers and grasping conditions. Further research into alternative feature space representations for temporal capacitive proximity sensing data, such as frequency domain analysis or advanced signal decomposition techniques, could promote greater improvements to classification and manipulation of varied media. Additional uses for capacitive sensing arrays could demand greater precision. 

\bibliographystyle{IEEEtran}
\bibliography{IEEEabrv, references}

\end{document}